\pdfoutput=1

\documentclass[11pt]{article}

\usepackage[final]{acl}

\usepackage{times}
\usepackage{latexsym}
\usepackage{booktabs}
\usepackage[T1]{fontenc}

\usepackage[utf8]{inputenc}

\usepackage{microtype}

\usepackage{inconsolata}

\usepackage{graphicx}

\definecolor{citecolor}{HTML}{0071BC}
\definecolor{linkcolor}{HTML}{D32F2F}%
\definecolor{cellcolor}{HTML}{E3F2FD}
\definecolor{red}{HTML}{D32F2F} %
\definecolor{greenish}{rgb}{0.757, 0.867, 0.004}
\definecolor{green}{HTML}{3CB371}  %
\definecolor{magenta}{HTML}{D81B60}
\definecolor{standardblue}{HTML}{0000FF}
\definecolor{standardred}{HTML}{FF8C00}
\definecolor{standardgreen}{HTML}{008000}

\usepackage{hyperref}
\usepackage{url}
\hypersetup{colorlinks=true, linkcolor=black, citecolor=citecolor,urlcolor=magenta}

\title{{\color{standardred}Fin}{\color{citecolor}LLM}-{\color{standardgreen}B}: When {\color{citecolor}L}arge {\color{citecolor}L}anguage {\color{citecolor}M}odels\\ Meet {\color{standardred}Fin}ancial {\color{standardgreen}B}reakout Trading}

\author{
  \textbf{Kang Zhang\textsuperscript{1,2}},
  \textbf{Osamu Yoshie\textsuperscript{2}},
  \textbf{Lichao Sun\textsuperscript{3}},
  \textbf{Weiran Huang\textsuperscript{1,4,}}\thanks{Correspondence to Weiran Huang. This work has been accepted to NAACL 2025 as an industry paper.}
\\[0.5em]
  \textsuperscript{1}MIFA Lab, Qing Yuan Research Institute, Shanghai Jiao Tong University, Shanghai, China
\\
  \textsuperscript{2}Waseda University, Tokyo, Japan
\quad
  \textsuperscript{3}Lehigh University, Bethlehem, PA, USA
  \\
  \textsuperscript{4}State Key Laboratory of General Artificial Intelligence, BIGAI, Beijing, China
\\[0.2em]
{zhangkang@toki.waseda.jp, yoshie@waseda.jp, lis221@lehigh.edu, weiran.huang@outlook.com}
  }

\usepackage{fancyhdr}
\begin{document}

\maketitle
\thispagestyle{fancy}
\fancyhf{}
\lhead{\includegraphics[height=1.6em]{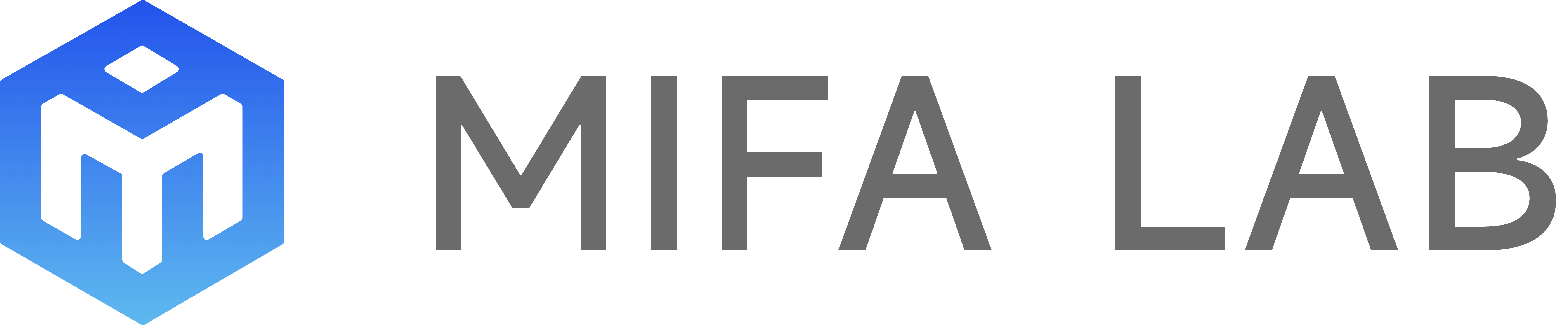}}
\renewcommand{\headrulewidth}{0pt}

\begin{abstract}
Trading range breakout is a key method in the technical analysis of financial trading, widely employed by traders in financial markets such as stocks, futures, and foreign exchange. However, distinguishing between true and false breakout and providing the correct rationale cause significant challenges to investors. Traditional quantitative methods require large amounts of data and cannot directly present the reasoning process, making them less than perfect in this field. Recently, large language models have achieved success in various downstream applications, but their effectiveness in the domain of financial breakout detection has been subpar. The reason is that the unique data and specific knowledge are required in breakout detection. To address these issues, we create the first financial breakout dataset and introduce FinLLM-B, the premier large language model for financial breakout detection, which enhances the effectiveness of breakout trading strategies. Furthermore, we have developed a novel framework for large language models, namely multi-stage structure, effectively reducing mistakes in downstream applications. Experimental results indicate that compared to GPT-3.5, FinLLM-B improves the average accuracy of answers and rational by 49.97\%, with the multi-stage structure contributing 9.72\% to the improvement. Additionally, it outperforms ChatGPT-4 by 42.38\%.

\end{abstract}

\hypersetup{colorlinks=true, linkcolor=linkcolor, citecolor=citecolor,urlcolor=magenta}
\section{Introduction}
Fundamental and technical analysis are the primary methods in financial investment. Given the limitations of the efficient market hypothesis in real financial markets~\cite{ball2009global,malkiel2003efficient,stout2002mechanisms}, the significance of technical analysis is recognized~\cite{blume1994market,taylor1992use,lo2000foundations,knight2010chart}. Trading range breakouts,
\begin{figure}[ht]
    \centering

    \includegraphics[trim={16.02cm 0.3cm 0.76cm 0.1cm},clip,width=1\linewidth]{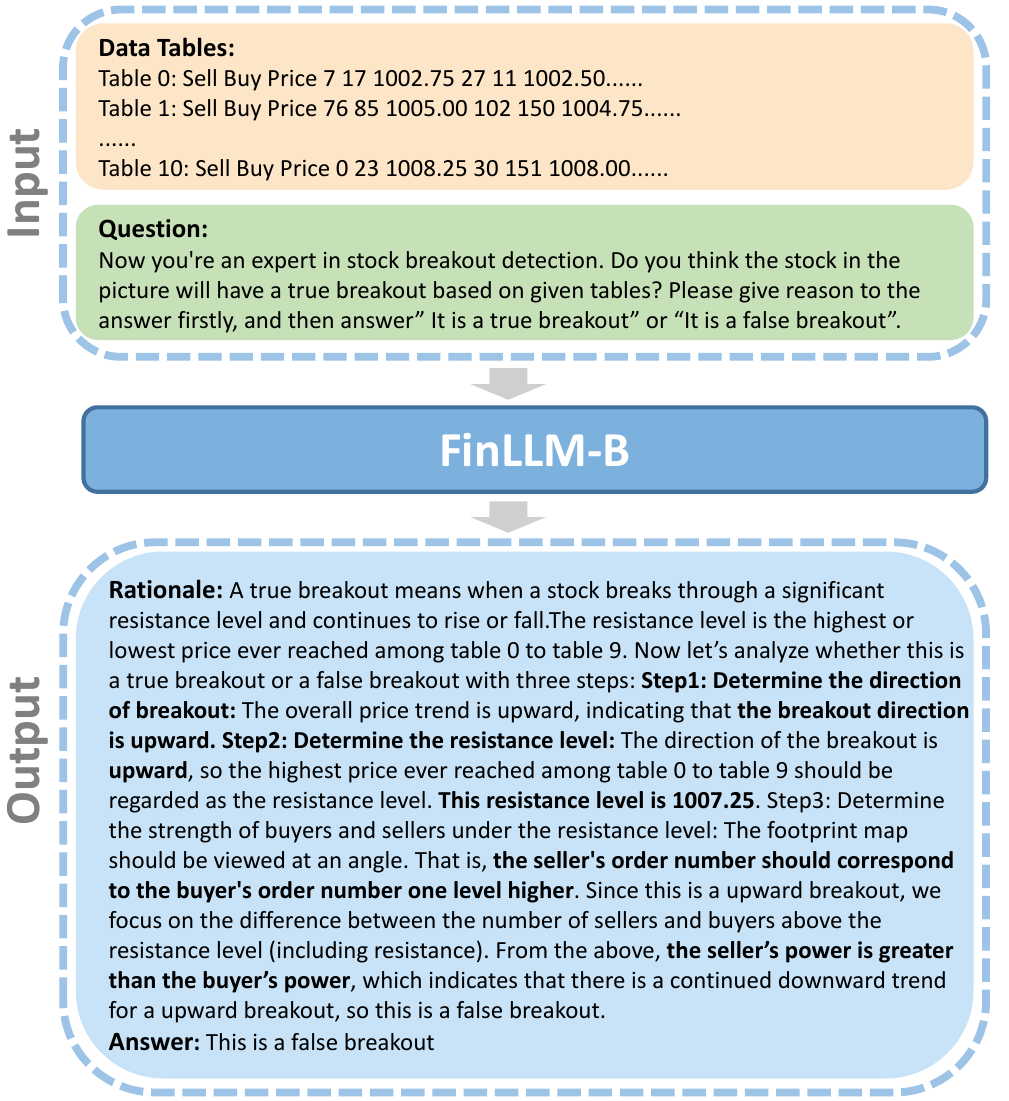}

    \caption{A sample of FinLLM-B task.}
    \label{fig:Sample}
\end{figure}
a key aspect of technical analysis, have been shown to be highly effective~\cite{zhu2015profitability,lubnau2014technical}. 
However, breakouts are often accompanied by false signals, which makes detecting false breakouts an important issue of breakout trading strategy~\cite{zhang2020deep,elder2002come}. Traditional quantitative methods~\cite{han2023challenge,john2023stock,zhang2023stock,kim2019forecasting} struggle with breakout detection due to limitations in dataset accessibility and report readability. For dataset accessibility, breakout detection requires footprint data, which is not readily available in mainstream datasets, hindering model training. For report readability, the finance sector demands high model explainability to ensure transparent decision-making~\cite{laux2024trustworthy,ben2021explainable,fritz2022financial}. Addressing these challenges is crucial for improving breakout detection methods.

Large language models (LLMs) have shown promise in fine-tuning with limited data~\cite{brown2020language,gao2020making} and generating comprehensive reports with 
rationale. These characteristics make LLMs strong candidates for breakout detection. However, three challenges remain. Firstly, LLMs lack domain knowledge, as observed in our experiments with GPT-3.5 and GPT-4, which struggled with breakout detection queries due to insufficient specialized datasets. Secondly, LLMs often produce outputs with mistakes~\cite{mcintosh2023culturally,lee2023mathematical,zhang2023siren}, including incorrect resistance levels and trend analysis. Thirdly, LLMs exhibit output inconsistency~\cite{chang2024survey,tan2023evaluation}, which can significantly impact model performance in financial domain.

In this work, we introduce FinLLM-B, a LLM for financial breakout detection as shown in Figure~\ref{fig:Sample}. FinLLM-B supplements the foundational
knowledge of GPT-3.5 in breakout detection and employs a multi-stage framework to mitigate errors and instability. This framework segments the rationale, allowing FinLLM-B to focus on subtasks, improving both accuracy and stability. Our experiments show that FinLLM-B outperforms GPT-3.5, achieving a 49.97\% improvement.

Our contributions can be summarized as follows: 1) We introduce FinLLM-B, the first large language model for financial breakout detection, which demonstrates domain knowledge and helps improve the reliability of breakout trading strategies. 2) Financial breakout dataset. We create the first dataset for financial breakouts, providing a valuable resource for future research in this area. 3) Multi-stage structure. We propose a multi-stage structure that segments the rationale, effectively reducing errors and enhancing stability for large language models in downstream tasks.

\section{Related Work}

\paragraph{Trading Range Breakout.} Technical analysis focuses on predicting financial market movements based on historical chart data~\cite{murphy1999technical}, demonstrating its profitability~\cite{taylor1992use, lo2000foundations}. A key method within technical analysis is the trading range breakout~\cite{raj1996effectiveness, lento2007profitability, bessembinder1995profitability}, which suggests that a price struggle occurs between buyers and sellers at resistance levels. Once the price surpasses this resistance level, it forms a strong support, preventing a short-term price reversal~\cite{brooks2011trading, chordia2002order, gosnell1996intraday}. 

\paragraph{Large Language Models.} Large language models (LLMs) have shown success across various applications~\cite{wu2023bloomberggpt, li2023druggpt, luo2022biogpt, bi2023oceangpt, kraljevic2021medgpt, sarrion2023implications, liu2023deid, liu2021finbert, li2024llava}. A challenge of applications is generation of incorrect answers. One solution related to this study is chain-of-thought (CoT)~\cite{wei2022chain} which prompts LLMs to reason before providing answers. Pioneering works involved manually designing examples to teach models reasoning, enabling more accurate responses~\cite{wei2022chain}. Subsequent research introduced approaches like zero-shot-CoT~\cite{kojima2022large} and auto-CoT~\cite{zhang2022automatic}, though CoT does not fully eliminate incorrect outputs, and researchers have explored incorporating new modalities~\cite{zhang2023multimodal, lu2022learn}.
\begin{figure*}[ht]
    \centering

    \includegraphics[trim={0.46cm 3.64cm 0.2cm 0.58cm},clip,width=1\linewidth]{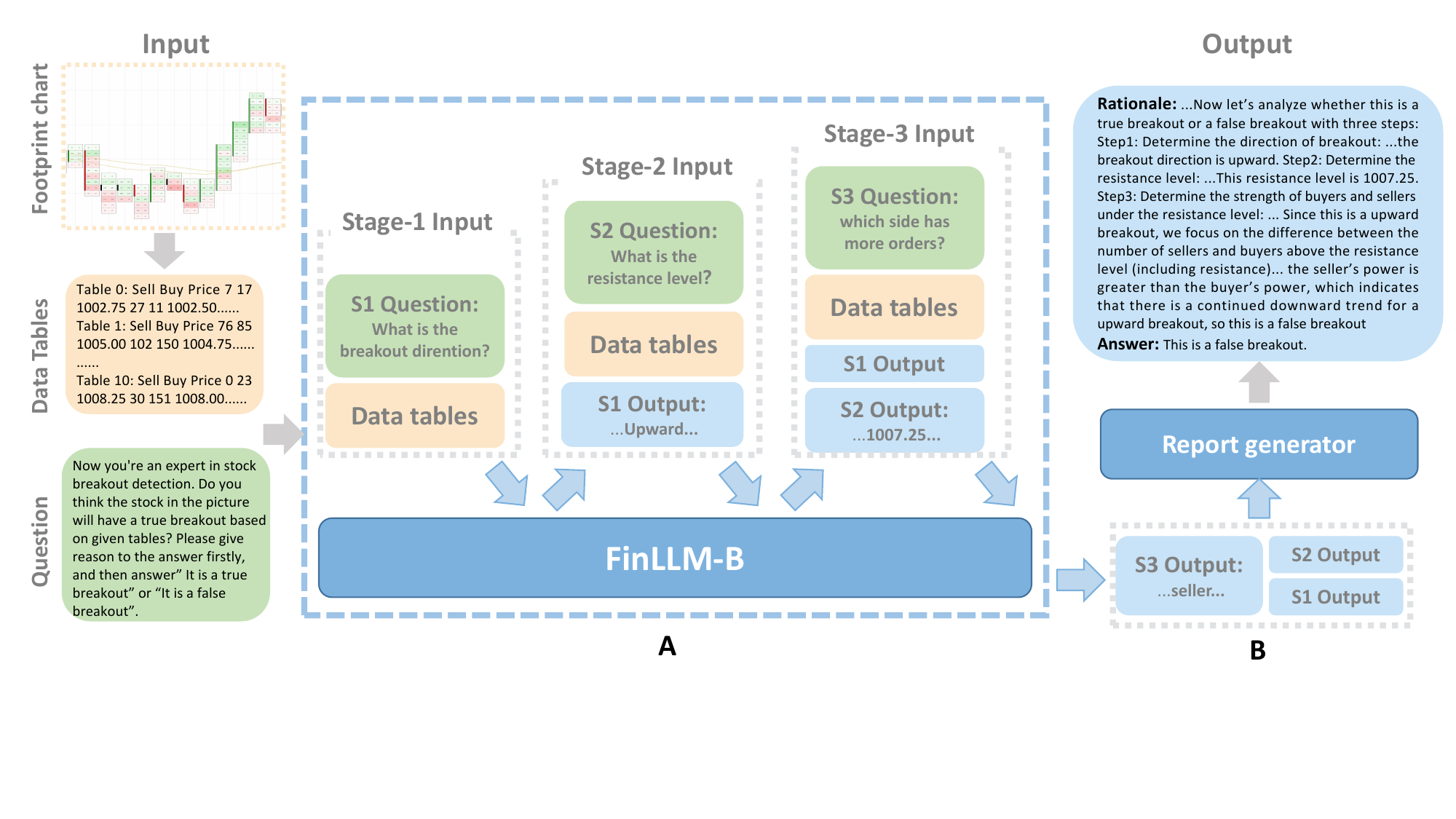}

    \caption{Overview of FinLLM-B with multi-stage structure. Multi-stage structure consists of two parts: Part A and Part B. Part A comprises three stages, each corresponding to a subtask of breakout detection. Part B is responsible for integrating the answers from Part A into a rationale and providing the final answer.}
    \label{fig:main}
\end{figure*}
\section{Problem Formulation}
Financial breakout detection is an important problem in the field of breakout trading. It determines whether a financial product is undergoing a true or false breakout, with true breakouts identified based on the order flow rule~\cite{valtos2015trading}. This study focuses on training a large language model to generate financial breakout detection reports with accurate rationales using processed data tables.

Time scale variability affects resistance levels and breakout authenticity, requiring a clear definition of the resistance level and true breakouts. The resistance level is defined as the highest or lowest price in the ten time ticks before the breakout~\cite{brooks2011trading,valtos2015trading}. A true breakout occurs when the closing price remains beyond the resistance level for two consecutive time units.

The primary input is a data table as shown in Figure~\ref{fig:Sample} derived from footprint charts. These charts capture detailed price information within each time unit, along with the order volumes from buyers and sellers at various price levels. Compared to historical stock line and candlestick charts, footprint charts offer richer detail, enabling more accurate assessments of breakout authenticity.

The output should include both the rationale and the answer as illustrated in Figure~\ref{fig:Sample}. This design is chosen because the investment field demands high explainability of decisions, and auditing the rationality behind decisions helps mitigate the risk of overvalued accuracy caused by guesses.

\section{Method}
Our model is designed for financial breakout detection, with inputs being prompts and specialized data tables. The multi-stage architecture is the main framework of our model, as shown in Figure~\ref{fig:main}. The reasons for its design are as follows. The amount of data is limited in our task. Researchers usually choose to tackle this task by fine-tuning models directly. In the initial trial of our study, we attempted to address the problem by directly fine-tuning with one LLM as well, but the results were unsatisfactory. We think that reasoning and drawing conclusions are the two main steps humans take to solve this task. Based on this, we create two distinct datasets and trained two LLMs respectively responsible for reasoning and conclusion: FinLLM-B and report generator. Under this structure, FinLLM-B focuses on the problem itself rather than the details of report generation.

However, simply splitting the whole model into two parts for FinLLM-B and the report generator still has limited improvement. We find that longer outputs tend to increase errors. Therefore, based on the steps to solve the problem, we divide the training set for FinLLM-B into three parts, each part responsible for answering one subtask with a standard answer. This design offers three advantages. Firstly, this structure provides a framework for breakout detection, serving as prior knowledge to compensate for the lack of data. Secondly, these sub-tasks have a sequential relationship. They share parameters and complement each other so that we can more effectively solve these subtasks with one large language model (FinLLM-B). Thirdly, each part answers only one question, allowing it to focus on specialized knowledge and provide concise responses. This approach is similar to the division of labor and cooperation within a human team, significantly enhancing the accuracy and stability of final outputs.

\subsection{Multi-Stage Structure}

The model consists of two parts: task flow (Part A) and report generator (Part B), as shown in Figure~\ref{fig:main}. 

\paragraph{Task Flow.} The task flow primarily consists of three parts: Stage 1 (S1) task, Stage 2 (S2) task, and Stage 3 (S3) task, which correspond to the three steps of breakout detection as follows. Firstly, we need to determine the direction of the entire breakout. If the historical price shows an upward trend, it indicates an upward breakout. Secondly, the resistance level of the breakout needs to be identified. Identifying the resistance level depends on the direction of the breakout. For an upward breakthrough, its resistance level is the historical price's highest value, defined as the highest price point in the ten time units preceding the current time. For a downward breakout, its resistance level is the historical price's lowest value. Thirdly, we need to compare the forces of buyers and sellers, with the comparison point varying based on the results of the previous two steps. For 
\begin{figure}[ht]
    \centering

    \includegraphics[trim={2.81cm 5.58cm 18.56cm 0.66cm},clip,width=1\linewidth]{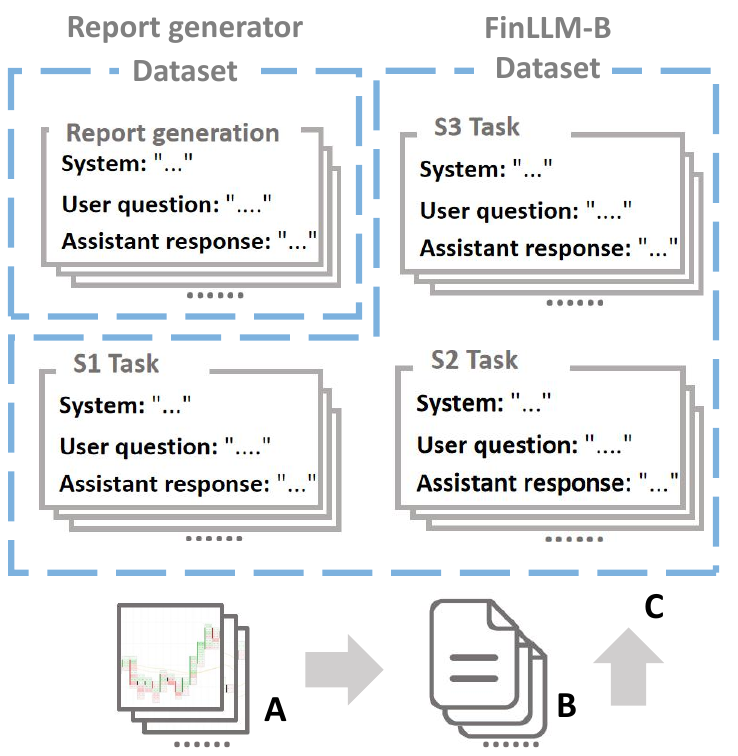}

    \caption{Dataset Construction. A: Footprint chart. B: Data table derived from the footprint chart. C: Dataset. It consists of two parts: FinLLM-B dataset and report generator dataset.}
    \label{fig:dataset}
\end{figure}
an upward breakout, we compare the number of buy and
sell orders above the resistance level, and vice versa for a downward breakout. The side with more orders is considered the stronger force.

FinLLM-B is employed to complete these three stages, providing evidence for breakout authenticity. It is pre-trained on GPT-3.5 and fine-tuned with 10 epochs for optimal performance.

\paragraph{Report Generator.} The Report Generator is another large language model in our study. Its function is to aggregate the answers from FinLLM-B in sub-tasks and output an analysis report with the conclusion on the authenticity of the breakout. It fundamentally differs from FinLLM-B in functionality, hence it is trained independently on GPT-3.5, focusing exclusively on report generation.

\subsection{Dataset}
The process of dataset construction is shown in the Figure~\ref{fig:dataset}.
The source data is collected as minute-level S\&P 500 future footprint data from the NinjaTrader platform. We convert the source data into a special data table and then build the dataset. Compared to getting raw data directly from the platform, this approach saves 90\% of the capital cost and provide better adaptability for LLMs. After obtaining the data tables, we use manual annotation to construct the dataset.The accuracy of human data annotation is ensured based on the expertise of the annotator and real market simulation.

\paragraph{FinLLM-B Dataset.}
This dataset involves two parts: training and testing. For training, The dataset consists of 60 training data. This includes 20 source data for each of S1, S2, and S3, and 10 samples for each of the true and false breakouts. For testing, the model will be tested a total of 1200 rounds, including 40 source data for each stage, and each test is repeated 10 times to test the stability of the model. Each training data consists of three parts based on the official setup of OpenAI: system, user question, and assistant response.  

\paragraph{Report Generator Dataset.}
The dataset of report generator is simpler because its task is not complex. It has 20 training data, which are annotated by experts according to the task, and other settings are consistent with FinLLM-B.

\section{Experiment}

\begin{figure*}[ht]
    \centering

    \includegraphics[trim={0.71cm 0.59cm 0.6cm 0.9cm},clip,width=1\linewidth]{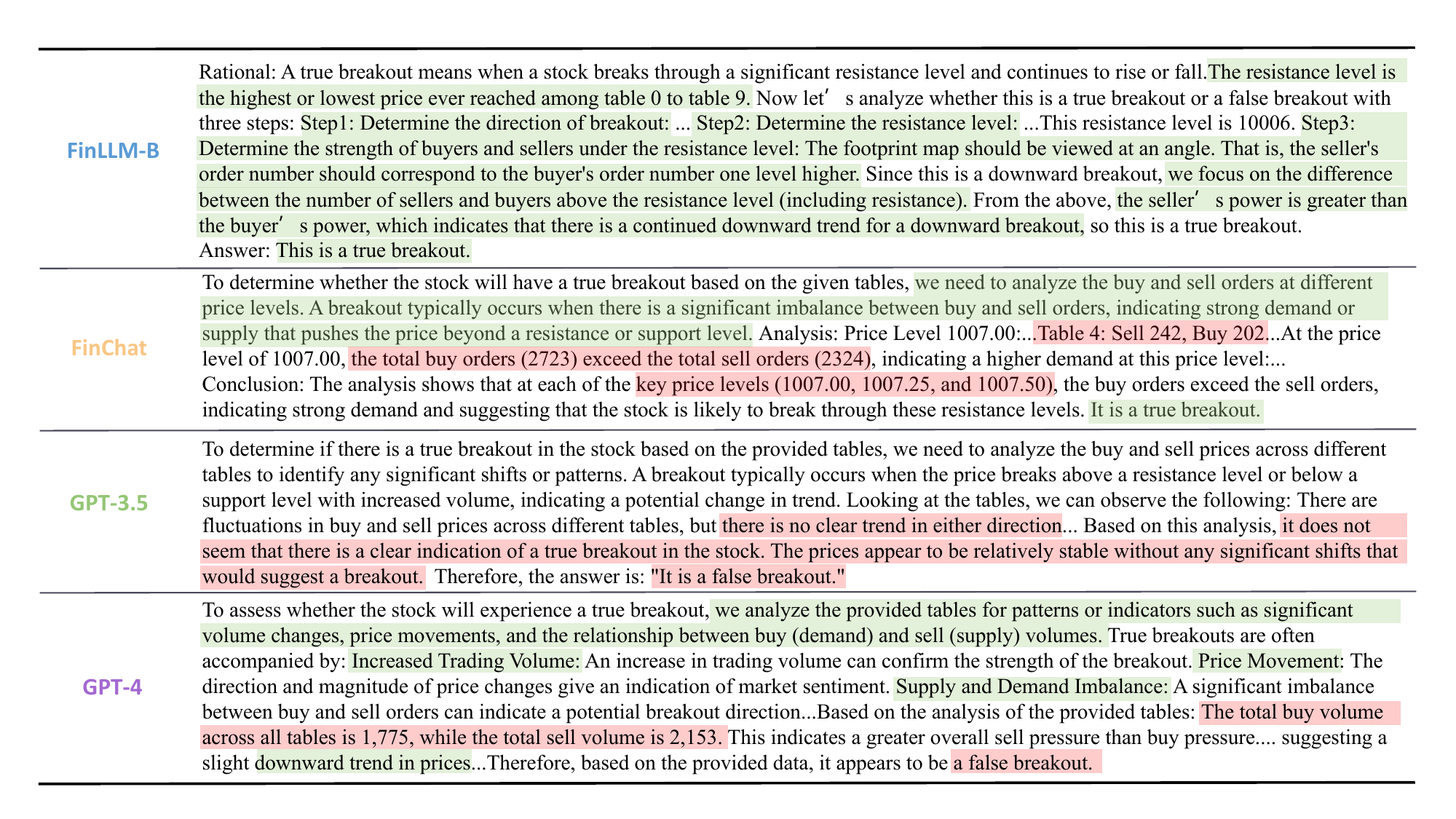}

    \caption{Output samples for professionalism comparison. Green: Valuable domain knowledge. Red: Incorrect domain knowledge and mistakes. Due to the length of the output, we used `...' to omit non-essential content.}
    \label{fig:model3}
\end{figure*}

\begin{table*}
\renewcommand{\arraystretch}{1.3}
\resizebox{0.95\textwidth}{!}{%
\begin{tabular}{c|ccc|c|c}
\toprule 
 Models &
  S1 Accuracy  &
  S2 Accuracy  &
  S3 Accuracy  &
  Average Accuracy  &
  Perfection Rate \\ \midrule
GPT-3.5 & $50.25 \pm 10.30$ & $10.50 \pm 5.99$ & $41.50 \pm 10.55$ & $34.83$ & $2.19$  \\
GPT-4  & $61.50 \pm 8.83$       & $13.50 \pm 4.74$      & $52.25 \pm 6.71$       & $42.42$ & $4.34$  \\
FinChat    & $75.5 \pm 8.96$   & $23.25 \pm 9.86$ & $60.50 \pm 5.99$  & $53.42$ & $11.18$ \\
LSTM       & --           & --          & --           & --     & $45$    \\ \midrule
FinLLM-B (Ours)  &
  $\mathbf{95.00} \pm \mathbf{0.00}$ &
  $\mathbf{89.40} \pm \mathbf{8.72}$ &
  $\mathbf{70.00} \pm \mathbf{0.00}$ &
  $\mathbf{84.80}$ &
  $\mathbf{59.45}$ \\ \bottomrule
\end{tabular}}
\caption{Result highlights. Accuracy and perfection rates of FinLLM-B and baseline models are evaluated based on correct identification of sub-tasks, reasoning process, and final breakout judgment. Note: LSTM only provides final results which are considered as the perfection rate.}
\label{Tab1}
\end{table*}

\subsection{Baseline \& Evaluation Metrics.}
FinLLM-B was trained based on GPT-3.5 and compared with four baselines: GPT-3.5~\cite{openai2022chatgpt}, GPT-4~\cite{achiam2023gpt}, FinChat~\cite{FinChat2024}, and Long Short-Term Memory network (LSTM)~\cite{bhandari2022predicting}. FinChat is a commercial-grade financial LLM that adapts GPT specifically for the finance sector. LSTM is a special recurrent neural network which is frequently used for financial prediction. We evaluated FinLLM-B in three main aspects: professionalism, accuracy, and stability.

\paragraph{Professionalism evaluation.} Since evaluating the expertise of the model's responses is subjective, we used manual scoring by professionals to assess the expertise of models.
\paragraph{Accuracy evaluation.} We compared the accuracy rate and perfection rate of each model. The accuracy rate is derived from the statistical analysis of the model's actual results. In addition, to evaluate the performance of the final report, we introduced the perfection rate, representing the proportion of samples that produced entirely accurate reports out of all test samples. An entirely accurate report correctly identifies each sub-task, the reasoning process, and the final breakout judgment. The calculation method is: S1 accuracy * S2 accuracy * S3 accuracy. Under this evaluation criterion, if the real market result is a true breakout, but the tested model's answer is a true breakout with incorrect reasoning, we consider the report inaccurate. This calculation method is necessary because having only the answers does not adequately reflect the model's capability.
\paragraph{Stability Evaluation.} Two testing methods were used to evaluate the stability of the model: standard deviation and output consistency distribution. For standard deviation, each model was tested 1200 rounds in total. We tested 40 sets of samples for task S1-3, each repeated 10 times and recorded each result for calculating the standard deviation. For output consistency distribution, we tested 40 sets of samples, each set tested 10 times repeatedly, and recorded the quantity of samples which produced same outputs across repeated tests. Specifically, we recorded the number of samples with 100\% same, 80\% same, 60\% same, and less than 60\% same. For example, if a test sample produces consistent outputs 8 times out of 10 repeated outputs, it is recorded as 80\% same in this round of testing. We are particularly concerned with cases where the outputs are 100\% same, indicating that the sample produced the same output all 10 times, demonstrating high reliability. We used the output consistency distribution because results of breakout detection will be used for investment decisions, thus requiring high consistency.

\subsection{Main Results} FinLLM-B outperforms other LLMs and neural network models, as shown in Table~\ref{Tab1}. It surpasses GPT-3.5 by 49.97\% in average accuracy and 57.26\% in perfection rate, primarily due to the baseline models' lower performance in the S2 task.

\subsection{Report Generator} We assessed the report generator's performance using 40 test samples, each tested 10 times. The generator consistently achieved expected results, due to the relatively simple nature of the task.
\begin{figure}
    \centering

    \includegraphics[trim={9.24cm 3.34cm 10.30cm 3.36cm},clip,width=0.9\linewidth]{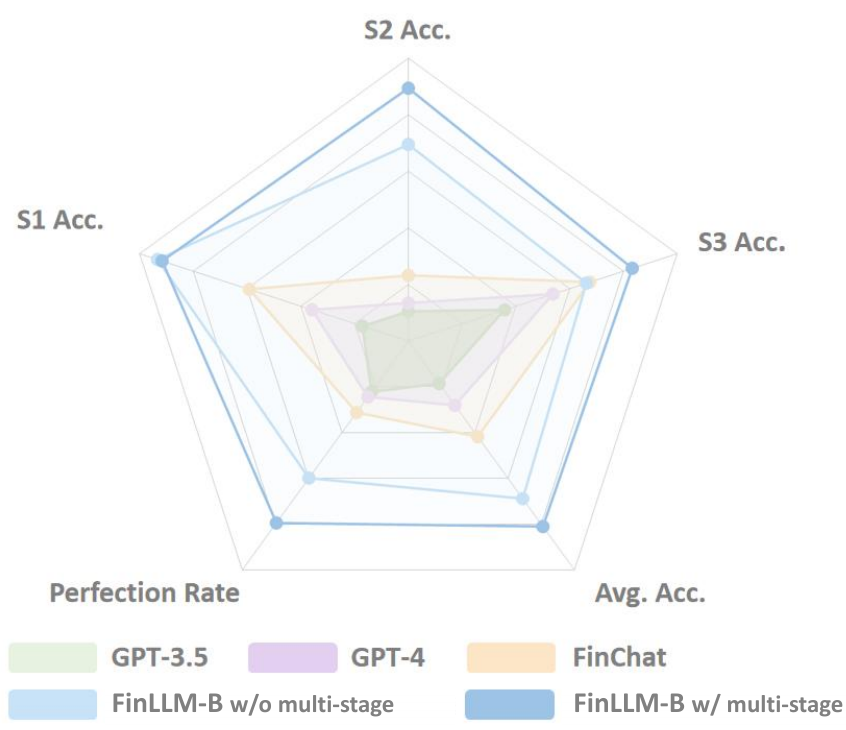}

    \caption{Accuracy comparison. Each axis is rescaled independently for better comparison.}
    \label{fig:rada}
\end{figure}

\begin{table}
\centering
\renewcommand{\arraystretch}{1.3}
\resizebox{0.9\linewidth}{!}{%
\begin{tabular}{ccc}
\toprule
FinLLM-B &  w/o multi-stage &  w/ multi-stage \\
\midrule
{S1 Accuracy} & {${96.00\pm 1.75}$} & $95.00\pm 0.00$ \\

{S2 Accuracy} & $69.50\pm5.63$ & ${89.40\pm 8.72}$ \\

{S3 Accuracy} & $59.75\pm 5.47$ & ${70.00\pm 0.00}$ \\
\midrule
{Average Accuracy} & $75.08$ & $\mathbf{84.80}$ \\

{Perfection Rate} & $39.87$ & $\mathbf{59.45}$ \\
\bottomrule
\end{tabular}}
\caption{Accuracy comparison between FinLLM-B with and without multi-stage. The proposed multi-stage structure demonstrates a notable improvement in the accuracy and perfection rate.}
\label{fig:table3}
\end{table}
\paragraph{Professionalism.} Scoring results reveal that FinLLM-B scored the highest, with an average of 8 out of 10, compared to GPT-4 and FinChat (6 out of 10) and GPT-3.5 (3 out of 10). Test samples shown in Figure~\ref{fig:model3} indicate that FinLLM-B demonstrates a clearer structure, more stable performance, and superior reasoning capabilities than the baselines.

\paragraph{Accuracy.} Figure~\ref{fig:rada} and Table~\ref{fig:table3} illustrates that FinLLM-B achieves significantly higher accuracy than other LLMs, especially in task S2. S2 task better highlights the model's strengths due to its uncountable answer space, unlike the countable answers in S1 and S2, where guessing inflates accuracy. LSTM's accuracy, close to 50\%, is limited by its requirement on substantial training data, which is difficult to obtain in our task.

\begin{figure}
    \centering

    \includegraphics[trim={1.45cm 14.45cm 3.81cm 5.93cm},clip,width=1\linewidth]{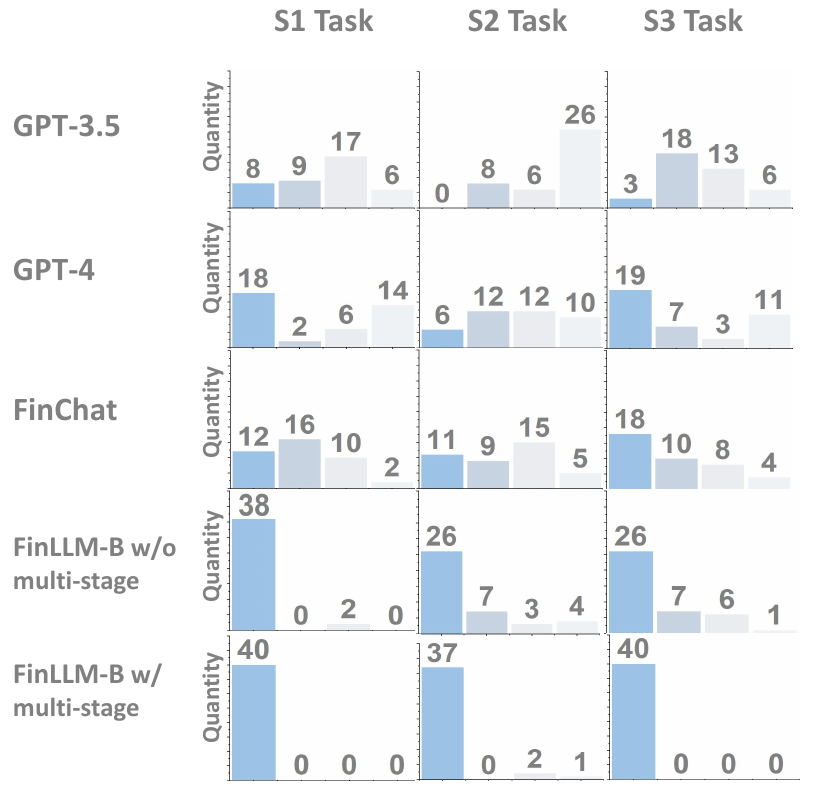}

    \caption{Output consistency distribution. Blue areas represent better stability. It represents the quantity of samples that have all same outputs in the stability test.}
    \label{fig:S_PIE100}
\end{figure}

\paragraph{Stability.} Figure~\ref{fig:S_PIE100} and Table~\ref{fig:table2} highlight FinLLM-B's stability advantages, particularly in S2. GPT-3.5's performance in S2 is significantly low. This is because the standard deviation here is the actual result's standard deviation, and GPT-3.5 often outputs values significantly different from the actual result. In Figure~\ref{fig:S_PIE100}, the blue area indicates the number of samples with all same output in 10 tests, demonstrating the stability of FinLLM-B. GPT-3.5 frequently switches between two answers, indicating that its accuracy is based on guessing.

\subsection{Report Generator} We assessed the report generator's performance using 40 test samples, each tested 10 times. The generator consistently achieved expected results, due to the relatively simple nature of the task.

\begin{table}
\centering
\renewcommand{\arraystretch}{1.3}
\resizebox{0.9\linewidth}{!}{%
\begin{tabular}{ccccc}
\toprule
 Models & S1 & S2 & S3 \\
\midrule
GPT-3.5 & $0.37$ & $170.05$ & $0.43$ \\
GPT-4 & $0.25$ & $0.39$ & $0.23$ \\
Finchart & $0.29$ & $0.32$ & $0.23$ \\
\midrule
FinLLM-B w/o multi-stage & $0.02$ & $0.14$ & $0.16$ \\
FinLLM-B w/ multi-stage & $\mathbf{0.00}$ & $\mathbf{0.06}$ & $\mathbf{0.00}$ \\
\bottomrule
\end{tabular}
}
\caption{Standard deviation. Actual resistance level values are used to calculate the standard deviation in S2.}
\label{fig:table2}
\end{table}

\subsection{Ablation Study} We compared FinLLM-B with and without the multi-stage structure, as shown in Figures~\ref{fig:rada}-\ref{fig:S_PIE100} and Tables~\ref{fig:table3}-\ref{fig:table2}. Two key findings emerged: 1) The multi-stage structure significantly improves accuracy, particularly in S2. 2) Stability is enhanced with the multi-stage design. These improvements arise from the structure's design. Under the multi-stage structure, the report generator handles report creation, allowing FinLLM-B to focus on answering questions. Each of three components in FinLLM-B specializes in a specific aspect, sharing parameters to enhance accuracy and stability.

\subsection{Dataset Size Analysis}
We tested the model's accuracy with different dataset sizes and found that the current 10 shots scale is appropriate. Samples in the dataset are categorized into two types: true and false breakout. We expanded the training set by increasing both the true and false breakout samples. For every 2 shots increase, we recorded the model's accuracy based on a single test run. The model accuracy for 2 to 10 shots is as follows: 57.50\%, 70.83\%, 78.21\%, 82.87\%, and 84.80\%. From the records, the rate of accuracy improvement slows down, and the rising trend curve becomes nearly flat at 10 shots. This indicates that our model's performance can improve with more training data, and at 10 shots, the performance is nearing its peak, suggesting that a 10-shot size is appropriate. Additionally, the model performed well with only 10 shots, further indicating that using LLMs is a promising approach for breakout detection in data-limited scenarios.

\section{Future Work}

Our work is the first to explore the application of large language models in financial breakout detection tasks, and we propose a multi-stage framework that enables our model to outperform other competitors. However, there is still room for improvement in the following two aspects.

Future work could expanding data modalities, such as images or videos, to better align the model with real-world scenarios. Currently, FinLLM-B relies on minute-level data from converted static footprint charts. However, the financial trading market changes rapidly, and continuous dynamic data could improve breakout detection accuracy. For instance, FinLLM-B could directly input videos to capture real-time changes in buy and sell orders in the future, enhancing breakout detection performance. Additionally, enriching the dataset with a broader range would provide deeper insights into the model’s optimal performance and robustness.

There is still room for improvement in the accuracy of the S3 task. We found the accuracy of S3 is significantly lower than the other two subtasks primarily due to its inherent complexity. The S3 task involves comparing the strength of buyers and sellers based on resistance levels, a process that is relatively intricate. This complexity may limit the full utilization of the capabilities of large language models. In the future, researchers could further segment the S3 task using a multi-stage structure to attempt to improve its accuracy.

\section{Conclusion}

We present FinLLM-B, the first large language model specifically designed for breakout detection, which alleviates the important issue in financial breakout trading field. To develop this model, we construct a high-quality financial breakout dataset. Furthermore, we create an innovative multi-stage framework, distinguishing FinLLM-B from the report generator and segmenting it into three distinct components based on problem-solving steps. This design enables FinLLM-B to more effectively demonstrate domain knowledge and enhances the model's accuracy and stability in our task. We believe that our model will serve as a valuable resource for future research and foster further exploration in the field of financial breakout trading.


\hypersetup{colorlinks=true, linkcolor=black, citecolor=citecolor,urlcolor=black}
\bibliography{custom}
\clearpage

\end{document}